\documentclass[conference,a4paper]{IEEEtran}
\IEEEoverridecommandlockouts
\usepackage{cite}
\usepackage{multirow}
\usepackage{amsmath,amssymb,amsfonts}
\usepackage{subcaption}
\usepackage{algorithmic}
\usepackage{graphicx}
\usepackage{textcomp}
\usepackage{xcolor}
\usepackage{siunitx}
\usepackage{tabu}

\def\BibTeX{{\rm B\kern-.05em{\sc i\kern-.025em b}\kern-.08em
    T\kern-.1667em\lower.7ex\hbox{E}\kern-.125emX}}
\begin{document}

\title{Accuracy evaluation  of a Low-Cost Differential Global Positioning System
	 for mobile robotics	\\
	
	\thanks{This work was partially supported by Bundesministerium fur Verkehr ¨
		und digitale infrastruktur (BMVI) in the context of the project 45KI02B021
		”Silicon Economy Logistics Ecosystem”}
}

\makeatletter 
\newcommand{\linebreakand}{%
\end{@IEEEauthorhalign}
\hfill\mbox{}\par
\mbox{}\hfill\begin{@IEEEauthorhalign}
}
\makeatother 


\DeclareRobustCommand*{\IEEEauthorrefmark}[1]{%
	\raisebox{0pt}[0pt][0pt]{\textsuperscript{\footnotesize #1}}%
}

\author{\IEEEauthorblockN{Christian Blesing\IEEEauthorrefmark{1}}
	\IEEEauthorblockA{\textit{christian.blesing@iml.fraunhofer.de}}
	\and
	\IEEEauthorblockN{Jan Finke\IEEEauthorrefmark{1}}
	\IEEEauthorblockA{\textit{jan.finke@iml.fraunhofer.de}}
	\and
	\IEEEauthorblockN{Sebastian Hoose\IEEEauthorrefmark{1}}
	\IEEEauthorblockA{\textit{sebastian.hoose@iml.fraunhofer.de}}
	\and
		
	\linebreakand

	\IEEEauthorblockN{Anneliese Schweigert\IEEEauthorrefmark{1}}
	\IEEEauthorblockA{\textit{anneliese.schweigert@iml.fraunhofer.de}}
	\and
	\IEEEauthorblockN{Jonas Stenzel\IEEEauthorrefmark{1}}
	\IEEEauthorblockA{\textit{jonas.stenzel@iml.fraunhofer.de}}
}

\maketitle

\footnotetext[1]{Fraunhofer	Institute for Materialflow and Logistic, Germany, Joseph-von-Fraunhofer-Straße 2-4, 44227 Dortmund}

\begin{abstract}
Differential GPS, commonly referred as DGPS, is a well-known and very accurate localization system for many outdoor applications in particular for mobile outdoor robotics. The most common drawback of DGPS systems are the high costs for both base station and receivers. In this paper, we present a setup that uses third-party open-source software and a Ublox ZED-F9P chip to build a ROS-enabled low-cost DGPS setup that is ready to use in a few hours. The main goal of this paper is to analyze and evaluate the repetitive and absolute accuracy of the system. The first measurement also examines the differences between a SAPOS \cite{rosenthal2001sapos} base station and a locally installed one consisting of low-cost components. During the evaluation process of the absolute accuracy, a moving mobile robot is used on the receiver side. It is tracked through a highly accurate VICON motion capture system  \cite{pfister2014comparative}. 
\end{abstract}


\section{Introduction}

DGPS approaches allow an accuracy improvement of GNSS positions, enabling more accurate outdoor localization of DGPS receivers than GPS receivers. DGPS receivers can significantly improve their GNSS positioning using correction signals provided by DGPS base stations \cite{farrell2000differential}. GPS receivers achieve accuracy that is in the range of several meters. DGPS receivers reach an accuracy of a few centimeters when using Real Time Kinematics (RTK) \cite{wang1999stochastic}.

\subsection{Related work}

DGPS localization approaches for outdoor vehicles and robots are being studied and improved by various researchers.
Many publications present approaches on how to improve raw DGPS localization data using probabilistic filtering techniques for the localization of moving vehicles \cite{hu2003adaptive} \cite{rohani2014dynamic} \cite{jo2015development}.
A few are focussing statical applications like monitoring of building structures with DGPS \cite{okiemute2018comparative}. In research applications, low-cost DGPS receivers from u-blox have been evaluated for many applications: Nitsch et al. \cite{nitsch2021embedded} are fusing GNSS and IMU measurements using an unscented Kalman Filter (UKF) leading to a positional accuracy of 1m. They compare the accuracy with data from a u-blox C099-F9P chip as a RTK pose reference.
In \cite{stranner2019high}, Stranner et al. present an accuracy comparison between a Novatel OEMv2 DGPS receiver and u-blox M8P-C94. 
Garcia et al. \cite{garcia2019accuracy} present an aproach for an image-based accuracy comparison between an eTrex 10 DGPS system and a u-blox NEO-6M receiver using landmarks on aerial images and a Trimble DGPS receiver as a reference. Hohensinn et al. \cite{hohensinn2022low} evaluate the accuracy of a similar DGPS setup as ours using different antennas using a KUKA industrial robot. Chosa et al. \cite{Chosa} evaluated the dynamic accuracy of a DGPS System using a turntable in motion with encoder increments as reference. 

\subsection{Contribution}
However, none of these publications has made a quantitative accuracy analysis using low-cost u-blox RTK-enabled base stations and receivers for static and dynamic localization applications using raw data. This is why we put a focus on a static experiment where we evaluate the repetitive accuracy of a u-blox ZED-F9P chip using a Small Load Carrier (SLC) with known dimensions. The second experiment evaluates the dynamic accuracy of the chip using a highly accurate VICON motion capture system as Ground-Truth (GT) position data. For both experiments we use available correction data from one of the 270 base stations of the German SAPOS network \cite{rosenthal2001sapos}.

\subsection{Content}

The subsequent sections of this paper are organized as follows. In section \ref{DGPS} the setup of the DGPS base station and receiver is described. Section \ref{repeatability} presents the experimental setup and the results of the static setting and in section \ref{absolute_accuracy}, the dynamic setting is described by its measurement setup and its results. Finally, in section \ref{conclusion} a conclusion is drawn that assesses the presented DGPS approach.

\section{System Overview}\label{DGPS}
In general, the implemented system can be divided into two main components, the base or reference station and the rover. First the base station including the used hard and software is described. Afterwards the mobile robot serving as the rover is explained in detail.

\subsection{Base Station}\label{base}

The base station, referred to as IML in the following, was set up as described here.
The base station is responsible for sending differential GPS correction data, also called RTCM data, over the internet. Taking a closer look at the base station, it consists of an active Ublox multi-band ANN-MB-00 antenna that is connected to a SparkFun GPS-RTK2 Board featuring the Ublox ZED-F9P \cite{hohensinn2022low}  module (see Figure \ref{fig:base}). To determine the coordinates of the base station antenna, the survey in mode from the ZED-F9P module was carried out for 62 hours. After these time period the determined antenna coordinates are applied and the ZED-F9P module was put into fixed mode to serve as a base station.

\subsection{Rover}\label{rover}

We use a highly dynamic omnidirectional mobile outdoor robot, called O³dyn (see Figure \ref{fig:absolute_accuracy_setup} and \cite{ullrich2022analyse}) for the evaluation of the absolute accuracy of our DGPS setup. This robot has the dimensions 2005 mm x 1450 mm x 744 mm (l, w, h). Two Ublox ANN-MB-00 antennas connected to two Ublox ZED-F9P modules are used for the experiments within this paper. All sensors, including both Ublox modules, are connected to the robot's control computer running ROS with an NTRIP client running as a ROS node. Within our evaluation we only use one of the modules thus neglecting the orientation of the robot.

\begin{figure}[hbtp!]
	\centering
	\includegraphics[width=\linewidth]{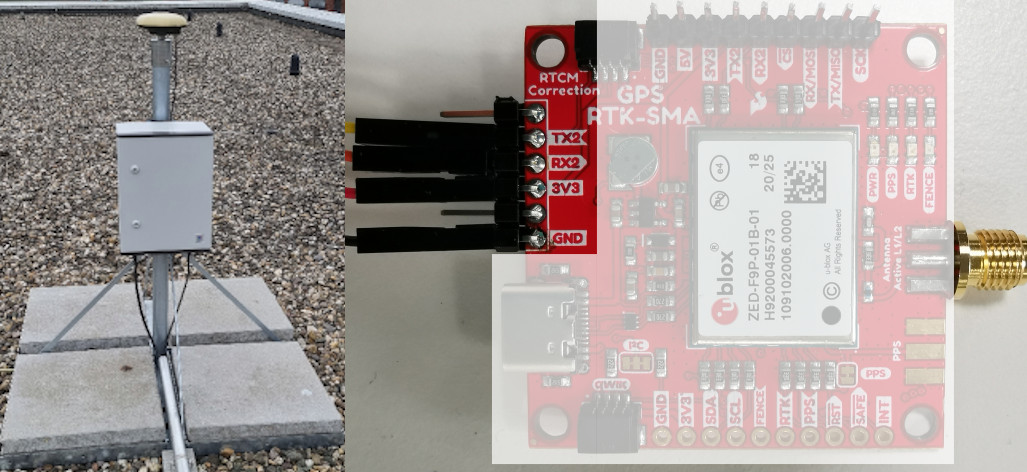}
	\caption{IML Base-Station with Ublox ZED-F9P}\label{fig:base}
\end{figure}

\section{Evaluation}\label{evaluation}

The evaluation is divided into two experiments with different measurement setups and results. The repetitive accuracy experiment is presented first and the absolute dynamic accuracy experiment is presented second.

\subsection{Repetitive Accuracy}\label{repeatability}

This section contains the static setting with the GT object being a SLC with the size of 0.596 m x 0.398 m x 0.22 m (length $\times$ width $\times$ height).

\subsubsection{Measurement Setup}

To evaluate the repeatability, the bottom of an SLC is facing upwards (see Figure \ref{fig:slc_tower}), so that a cross-shaped marking can be applied to the top of the SLC, which consists of two lines perpendicular to each other. The start and end points of these two lines are breakpoints for the GNSS antenna.
A fixture is placed on the edge of the SLC to which a GNSS antenna is attached. The fixture with the GNSS antenna is slid along the four edges of the SLC, stopping at the four breakpoints P0-P3 for 10 seconds each. The fixture with the GNSS antenna is slid around the SLC for a total of five rounds. The experiment starts at breakpoint P0 and the mount is manually slid to breakpoint P1 within 20 seconds. This procedure is repeated until breakpoint P0 is reached again, which starts a new round. 

\begin{figure}[hbtp!]
	\centering
	\includegraphics[width=0.7\linewidth]{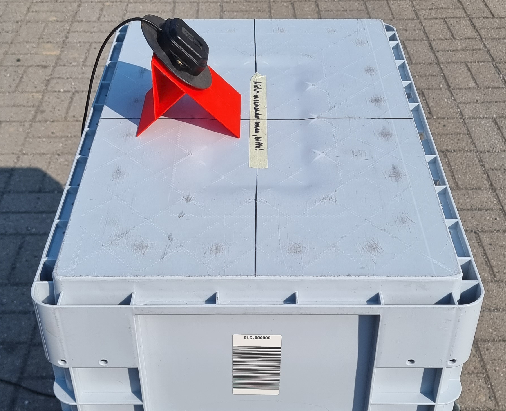}
	\caption{The measurement setup for the repeatability experiment}\label{fig:slc_tower}
\end{figure}

Measurements are taken for 8 seconds at each breakpoint. The receiver produces five measurements each second. When the antenna moves, measurements are also expected which follow the contours of the SLC.

\begin{figure}[htbp!]
	\centering
	\includegraphics[width=\linewidth]{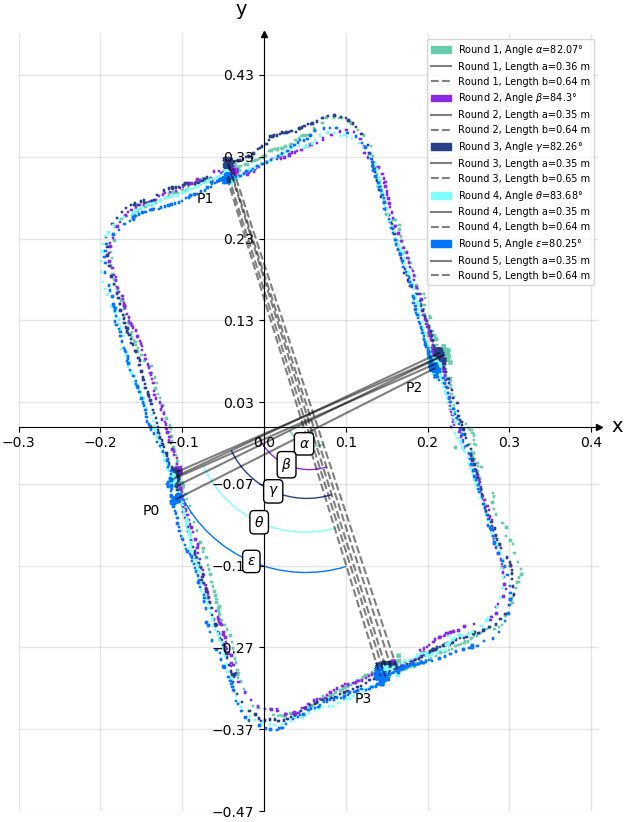}
	\caption{The measurements with the U-blox chip and SAPOS correction data are arranged over five rounds.}\label{fig:rep_img}
\end{figure}

\subsubsection{Results}

First, it is evaluated whether the dimensions of the GT object match the measurements of the antenna where the antenna is not in motion (see Figure \ref{fig:rep_img}). The relevant GT information is the angle $\alpha_{GT}$ between the perpendicular lines, the lengths $a_{GT}$ and $b_{GT}$ of a SLC, since the length between the two midpoints opposite and the angle between the two resuting lines are determined from P0-P3. The results between the SAPOS base station and a self-built base station(see Table \ref{tab:repeatability_results}) differ over all rounds only in the millimeter range and by less than one degree. Therefore, both base stations provide a similarly good correction of the GPS signal to map the lenghts, as well as the rectangular shape of the SLC.

\begin{table}[!htbp]
	\centering
	\caption{The results with the Ublox chip and the IML and SAPOS correction data are each averaged over five rounds.}
	\label{tab:repeatability_results}
	\begin{tabu}{l|l|l|l}
		
		Base Station & Avg. angle [°] & Avg. length $a$ [m] & Avg. length $b$ [m] \\ \tabucline[1.5pt]{-}
		
		IML          & 82.12      & 0.356         & 0.642         \\ \hline
		SAPOS        & 82.512     & 0.352         & 0.644         \\ 
	\end{tabu}
\end{table}

Then it is evaluated how well the physical breakpoints P0-P3 are obtained by corrected measurements respectively from the IML and SAPOS base station. Table \ref{tab:iml_sapos_breakpt_results} shows the result of the measurement. Both stations manage to correct the position similarly well, although the standard deviation and span are higher for the correction by the IML station.


\makeatletter
\newcommand{\thickhline}{%
	\noalign {\ifnum 0=`}\fi \hrule height 3pt
	\futurelet \reserved@a \@xhline
}
\newcolumntype{"}{@{\hskip\tabcolsep\vrule width 1pt\hskip\tabcolsep}}
\makeatother

\begin{table}[!htbp]
	\centering
	        \setlength\tabcolsep{2pt}
		\caption{Repeatability accuracy results for the points P0-P3 from the Ublox chip and correction data from both Base Stations}
		\label{tab:iml_sapos_breakpt_results}
	\begin{tabu}{c|[1.5pt]c|l|l|l|[1.5pt]l|l|l|l}
	
		& \multicolumn{4}{c|[1.5pt]}{SAPOS}        & \multicolumn{4}{c}{IML Base-Station}          \\ \tabucline[1.5pt]{-}
		matric & P0 [m]   & P1 [m]  & P2 [m]   & P3 [m]  & P0 [m]  & P1 [m] & P2 [m] & P3 [m]   \\ \tabucline[1.5pt]{-}
		
		x-mean & -0.107 & -0.043 & +0.213 & +0.151  & -0.005 & +0.067 & +0.317 & +0.255   \\ \cline{0-8} 
		y mean & -0.068 & +0.313  & +0.082 & -0.298 & +0.012  & +0.394 & +0.161 & -0.222  \\ \cline{0-8} 
		x std  & +0.002  & +0.002  & +0.003 & +0.003  & +0.002  & +0.003 & +0.004 & +0.004   \\ \cline{0-8} 
		y std  & +0.003  & +0.003  & +0.004 & +0.004  & +0.003  & +0.005 & +0.008 & +0.007   \\ \cline{0-8} 
		x min  & -0.110 & -0.046 & +0.206 & +0.146  & -0.010  & +0.063 & +0.309 & +0.246   \\ \cline{0-8} 
		x max  & -0.104 & -0.038 & +0.219 & +0.163  & -0.001 & +0.075 & +0.326 & +0.263   \\ \cline{0-8} 
		y min  & -0.073 & +0.305  & +0.070 & -0.304 & +0.007  & +0.381 & +0.144 & -0.238  \\ \cline{0-8}  
		y max  & -0.063 & +0.318  & +0.090 & -0.289 & +0.020   & +0.404 & +0.177 & -0.207  \\ \cline{0-8} 
		x span & +0.006  & +0.009  & +0.013 & +0.018  & +0.009  & +0.012 & +0.017 & +0.017   \\ \cline{0-8} 
		y span & +0.010  & +0.013  & +0.019 & +0.015  & +0.010   & +0.023 & +0.032 & +0.031  	\\ \cline{0-8}
	\end{tabu}

\end{table}

\subsection{Absolute Accuracy dynamic movement}\label{absolute_accuracy}

This section contains the the dynamic accuracy measurement using O³dyn and the VICON Ground-Truth.

\subsubsection{Measurement Setup}

The absolute accuracy of the system was evaluated using a motion capturing system as a GT reference  (see Figure \ref{fig:absolute_accuracy_setup}). A coverage of $\SI{60}{\metre\squared}$ was achieved with 12 cameras at a frame rate of $\SI{120}{\hertz}$.  DGPS coordinates habe been received with $\SI{5}{\hertz}$. All systems were time synchronized using Chrony  \cite{soares2020analysis}. During the dynamic measurements, the O³dyn passed through the observed area with defined velocities. GPS coordinates and GT information were logged to a ROS Bag.
The origins of both systems have been aligned by taking the GPS coordinates of the motion capturing center marker as the local origin of the cartesian coordinate system. A modified evo-Libary \cite{grupp2017evo} was used for the alignment and evaluation. Alignment base on \cite{Umeyama1991LeastSquaresEO} has been applied without scaling. Light conditions during the measurements allowed a reliable tracking up to a velocity of $\SI{4.12}{\meter/\second}$ .

\begin{figure}[hbtp!]
	\centering
	\includegraphics[width=\linewidth]{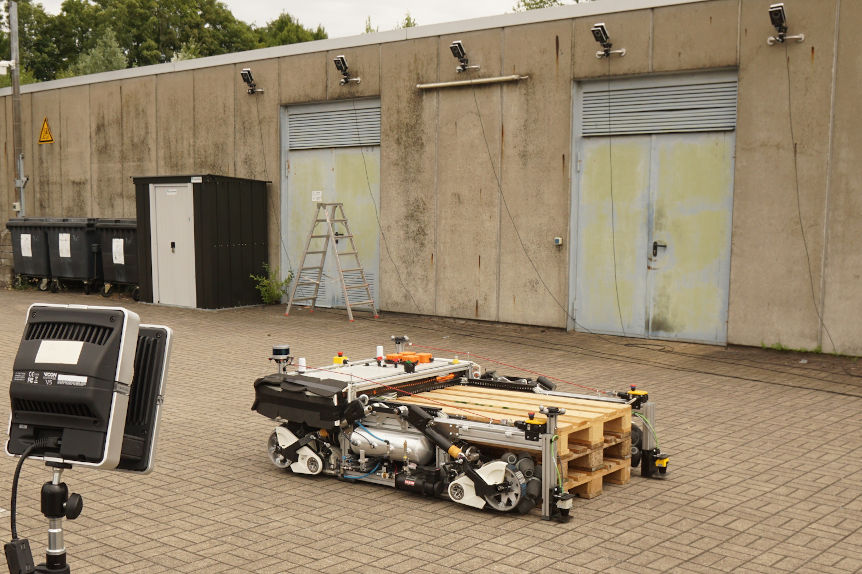}
	\caption{Measurement setup for the dynamic accuracy experiment with local origin at Lat=51.49457\textdegree\ Long=7.40745\textdegree}\label{fig:absolute_accuracy_setup}
\end{figure}

\subsubsection{Results}


\begin{table}[!htbp]
	\centering
	\caption{Accuracy results of O³dyn moving at different velocities}
	\label{tab:accuracy_sopas}
	\begin{tabu}{l|[1.5pt]l|l|l|l|l}
		
		& \multicolumn{4}{c}{Pose Translation-Error [m] }&  \\ \tabucline[1.5pt]{-}
		
		Mean velocity 		 						& max 		& mean 		& median 	& min 		& rmse          \\  \tabucline[1.5pt]{-}
		
		$\SI{1.10}{\meter/\second}$               	& 0.1058   	& 0.0691    & 0.0727   	& 0.0229	& 0.0728            \\  \hline
		$\SI{1.68}{\meter/\second}$            		& 0.1517   	& 0.1112    & 0.1165  	& 0.0257    & 0.1137            \\  \hline
		$\SI{2.07}{\meter/\second}$               	& 0.1839   	& 0.1110    & 0.1415   	& 0.0070    & 0.1254            \\  \hline
		$\SI{2.57}{\meter/\second}$             	& 0.2385   	& 0.1219    & 0.1705   	& 0.0072    & 0.1467           \\   \hline
		$\SI{3.12}{\meter/\second}$               	& 0.3120   	& 0.1508    & 0.1870   	& 0.0067    & 0.1751            \\  \hline
		$\SI{4.12}{\meter/\second}$                	& 0.4167   	& 0.2258    & 0.2400   	& 0.0438    & 0.0438            \\  \hline
		
	\end{tabu}
\end{table}

The Table  \ref{tab:accuracy_sopas} shows the results for the measurements of the dynamic accuracy. We can see that the accuracy is in the cm range. 

\section{Conclusion}\label{conclusion}

Our experiments show a good repetitive accuracy using low-cost receivers and correction data from a locally installed base station. Both stations provide similar  results within the $\SI{}{\centi\meter}$ range. The dynamic measurements of the absolute accuracy provide significantly less accurate results since the receiver is in motion. These observations are consistent with the measurements from \cite{Chosa}. When looking at these results it is important to keep in mind that the transformation is done by means of a least-squares estimation approach. 

\section{Outlook}\label{outlook}

For further experiments, it would be desirable to observe a larger measurement area and fuse the results from two antennas to investigate the robot orientation. It would also be interesting to compare the low cost system with a professional-grade one. 



\clearpage

\bibliographystyle{IEEEtran}
\bibliography{literature.bib}

\end{document}